\documentclass[conference]{IEEEtran}
\IEEEoverridecommandlockouts

\usepackage[cmex10]{amsmath}
\usepackage[caption=false]{subfig}
\usepackage{lipsum}
\usepackage[utf8]{inputenc}
\usepackage[mathscr]{euscript}
\usepackage{mathbbol}
\usepackage{xcolor}
\usepackage{blindtext}
\usepackage{breqn}
\usepackage{balance}
\usepackage{color}
\usepackage{multirow}
\usepackage{cite}
\usepackage{enumitem}
\usepackage[]{algorithm2e}
\usepackage{url}
\usepackage{comment}
\usepackage{enumitem}

\usepackage{amsmath,amssymb,amsfonts}
\usepackage{algorithmic}
\usepackage{graphicx}
\usepackage{textcomp}
\usepackage{xcolor}
\usepackage{color}
\def\BibTeX{{\rm B\kern-.05em{\sc i\kern-.025em b}\kern-.08em
    T\kern-.1667em\lower.7ex\hbox{E}\kern-.125emX}}
\begin{document}

\title{Reinforcement Learning for Battery Energy Storage Dispatch augmented with Model-based Optimizer}

    \author{Gayathri Krishnamoorthy,~\IEEEmembership{Student Member,~IEEE,}
    Anamika Dubey,~\IEEEmembership{Member,~IEEE,} and  \\
    Assefaw H. Gebremedhin, {Member,~IEEE}
        \thanks{All authors are with the School of Electrical Engineering and Computer Science, Washington State University, Pullman, WA, 99164 e-mail: g.krishnamoorthy@wsu.edu, anamika.dubey@wsu.edu, assefaw.gebremedhin@wsu.edu.}
    }

\maketitle
\begin{abstract}
Reinforcement learning has been found useful in solving optimal power flow (OPF) problems in electric power distribution systems. However, the use of largely model-free reinforcement learning algorithms that completely ignore the physics-based modeling of the power grid compromises the optimizer performance and poses scalability challenges. This paper proposes a novel approach to synergistically combine the physics-based models with learning-based algorithms using imitation learning to solve distribution-level OPF problems. Specifically, we propose imitation learning based improvements in deep reinforcement learning (DRL) methods to solve the OPF problem for a specific case of battery storage dispatch in the power distribution systems. The proposed imitation learning algorithm uses the approximate optimal solutions obtained from a linearized model-based OPF solver to provide a good initial policy for the DRL algorithms while improving the training efficiency. The effectiveness of the proposed approach is demonstrated using IEEE 34-bus and 123-bus distribution feeders with numerous distribution-level battery storage systems.
\end{abstract}


\section{Introduction}

With the integration of controllable technologies at the grid-edge such as smart inverters and battery energy storage systems (BESS), the use of optimal power flow (OPF) algorithms has become pervasive in power distribution systems \cite{VV01}. 
Traditional optimization algorithms used to solve nonlinear OPF problems pose convergence and scalability challenges {\cite{VVC1,VV05}}. More importantly, the use of mathematical optimization techniques is challenged for the application cases that require a time-sensitive response \cite{RL3}. One such application is the dispatch of distribution-level BESS for frequency control applications as per the FERC orders 841 and 2222 \cite {ferc841,ferc2222}. The existing mechanisms that employ area control error (ACE) calculations~\cite{ref12}, do not account for the impacts of dispatching distribution-connected BESS on power distribution systems \cite{ref11}. With growing applications for distribution-connected assets in providing demand-side flexibility, the satisfaction of distribution-level operating constraints is imperative when provisioning bulk grid services from distribution-connected assets \cite{Johanna1}. Traditionally, a distribution-level BESS dispatch problem can be formulated as an OPF. However, given the fast response needed for battery dispatch in frequency control applications, the use of mathematical optimization approaches is inadequate as they are slow and do not scale well. Another major limitation of optimization-based approaches is the need for the state information for the entire distribution system, which is cost-prohibitive \cite{RL4}. Within this context, this paper addresses the problem of incorporating a large number of distributed and controllable BESS for frequency control applications with consideration to distribution-level operating constraints.


Given the challenges of purely model-based optimization methods, data-driven model-free reinforcement learning (RL) approaches have recently emerged as an attractive alternative to solving distribution-level OPF problems. A detailed survey summarizing the applications of RL methods in power grid operation and control is provided in \cite{RL1, RL2}. Also refer to recent work on the applications of RL algorithms for voltage control applications \cite{RL6, RL7,RL10, RL11}. Unfortunately,  existing model-free RL algorithms ignore the crucial information embedded in the physics-based model of the power distribution systems and may thus compromise the optimizer performance and pose scalability challenges. In more recent works, including power systems model information in neural networks has shown to improve the performance of the OPF problems \cite{RL8, RL9}. 

This work proposes a novel use of imitation learning (IL) methods to synergistically combine physics-based model information with learning-based control for fast real-time control of distribution-connected BESS with consideration to distribution-level operating constraints. First, we introduce a constrained MDP formulation for the optimal BESS dispatch problem with distribution-level operating constraints. A model-free constrained soft-actor critic (CSAC) algorithm is used to solve the resulting constrained MDP. Next, we augment the training of the DRL agent with a model-based optimizer using IL-based improvements. In the machine learning community, IL methods have been extensively used to include the expert decisions (typically human agents) to train the RL agents \cite{IL2}. Such formalisms aid learning by providing agents examples to imitate. 
Thus, we propose to use an approximate model-based optimizer of low-compute complexity (using linear power flow equations) to generate near-optimal solution trajectories that are then used as expert demonstration trajectories to augment the training of the DRL agent (with CSAC algorithm) using Soft-Q Imitation Learning (CSAC-SQIL) \cite{sqil}. The proposed IL-based improvements not only provide a good initial policy for the DRL algorithms but also improve the training efficiency. 
It is shown that the proposed learning-based controllers result in near optimal solutions and are able to satisfy the power flow constraints. 

\section{Problem Formulation}
In this section, we detail the mathematical formulation for optimal dispatch of BESS for frequency control applications.
The problem formulation details the approach to optimally dispatch BESS in response to ACE regulation signal while maintaining critical device-level (BESS) and distribution-level operating constraints. First, we detail a centralized OPF problem to optimally dispatch the ACE regulation signals to the participating distributed BESS (Section II.A). Next, we introduce a combined DRL-IL approach to perform optimal BESS dispatch in response to the ACE signal (Section II.B). 


\subsection{Centralized OPF Formulation for BESS Dispatch}

At the distribution level, we assume a distribution network operator (DNO) acts as the DER aggregator and is responsible for optimally dispatching the dispersed BESS at the distribution level in response to the ACE signal received from the transmission system operator (TSO). Note that the ACE signal defines the required active power from the aggregated BESS at the DNO level. The generation of ACE signal is not within the scope of this work; readers are referred to the following related literature \cite{ref14}. The DNO's objective is to optimally dispatch the dispersed BESS to track the required active power demand from TSO (in the form of ACE signal) without violating the distribution network operational constraints, especially nodal voltage constraints under high levels of DER penetrations. We assume a linearized approximation for the distribution power flow model. The resulting optimization problem for the DNO is thus formulated as a mixed-integer linear programming (MILP) problem that can be solved using off-the-shelf optimization solvers. The centralized optimization problem for the DNO is formulated as follows. The variables in the formulation are defined below.
\begin{equation}
\small
\begin{aligned}
 \underset{P_{i,t}}{\text{Maximize}}
& \sum_{i=1}^m \sum_{t=1}^h |P_{i,t}|. Pr(RA_{i}) + \mathscr{P} (Cr,b) \delta C\\
\end{aligned}
\end{equation}

 \vspace{-0.6cm}
 
 \begin{small}
\begin{flushleft}
\begin{eqnarray}
\nonumber \text{subject to} \\
& \sum_{i=1}^m |P_{i,t}| \in [P_{target} - \epsilon_{M}, P_{target} + \epsilon_{M}] & \;  \\
& P_{i,t} \in [AI.\underbar{P} , AI. \overline{P}] & \;  \\
& 0 \leq \sum_{t=1}^h \lambda_{i} |P_{i,t}| d_{sch} \leq |RA_{i} . \Delta_{tot}| & \;  \\
& \mathscr{P}_{i} (Cr,b) \geq \rho^{min} &\; \\
& C \in [0, B] &\; \\
\nonumber & constraints  \hspace{0.1cm} (7)-(14)
\end{eqnarray}
\end{flushleft}
\end{small}

\noindent $t$= timestep  \\
$m$ = number of available resources \\
$\eta$ = single-trip charge or discharge efficiency \\
$P $ =  power level of resource $i$ during time $t$ \\
$\mathcal{P} $ =  power rating of resource $i$  \\
$\mathscr{P(.)} $ =  payment received by the regulation resource \\
$E$ = BESS energy in kWh \\
$C$ = regulation capacity in kW \\
$r$ = set of all normalized regulation instructions \\
$b$ = set of all battery dispatch power \\
$\delta$ = regulation market clearing price in \$/kW \\
$d_{sch}$ = discretized interval for optimization \\
$h$ = $(\bar{t} - \underbar{t})/d_{sched}$ = overall time horizon \\
$P_{target}$ = Requested power by TSO for frequency regulation \\
$ \epsilon_{M}$ = Dispatch tolerance of regulation market \\
$AI$ = Availability interval \\
$RA$ = Resource availability statement \\
$\Delta$ = change in energy storage level \\
$ \lambda_{i}$ = $\eta$. $C^{in}_{i}$ for charging \\
$ \lambda_{i}$ =  $C^{out}_{i}$ / $\eta$ for discharging 

\vspace{0.1cm}
The first term of the objective function aims at determining  feasible controls for achieving the target power level, $P_{target}$, specified by the TSO for frequency regulation in the form of ACE signal. The term $Pr (RA_{i})$ will prioritize the use of BESS resources. The second part maximizes the payment the DNO receives from the regulation market while accounting for its performance in the previous interval (performance index). Constraint (2) ensures that the resulting power level over the entire time does not deviate by more than $ \epsilon_{M}$ from the target power level $P_{target}$. Constraint (3) defines the admissible ranges for the decision variables. Constraint (4) ensures that no resource supplies or consumes more energy than its permissible limits. Constraints (5) and (6) enforce a minimum performance index $\rho^{min}$ to the response while limiting the regulation capacity to be within the rated battery capacity. The payment received by the participant in this market is $ \mathscr{P}_{i} (Cr,b) \delta C$, where $b$ is the participant response to the regulation instruction and $P(.)$ is the performance index calculation function. For simulation purposes, we consider default market regulation parameters from \cite {BES3}. In this work, we calculate the performance index as 
\begin{equation}
\small
    \mathscr{P}_{i} (Cr,b)  =1- \frac{|Cr-b|}{C|r|} \delta
\end{equation}

We detail the constraints for the battery model, (8)-(10).
\begin{equation}
\small
    \mathcal{P} \leq P_{i,t} \leq \mathcal{P}   
    \end{equation}
    \vspace{-0.6cm}
    \begin{equation}
      \small
        \underbar{E} \leq e_{i,t} \leq \overline{E} 
    \end{equation}
\vspace{-0.4cm}
\begin{equation}
\small
    e_{i,t} - e_{i,t-1} = d_{sch} \eta [P_{i,t}]^+ - d_{sch} [-P_{i,t}]^+ / \eta  
\end{equation}

\noindent where $P_{i,t} $ is the  battery dispatch power at time $t$ and $e_{i,t}$ is the energy level at step $t$. Equations (3), (4), and (5) model BESS power rating, energy rating, and the evolution of the battery state-of-charge, respectively. 

Finally, we formulate the operational model for the distribution system to be included in the DNO's battery dispatch problem (11)-(14). The distribution system is modelled as a directed graph $\mathcal{G}=(\mathcal{N},\mathcal{E})$, where  $\mathcal{N}$ and $\mathcal{E}$ denote the set of nodes and edges. Two nodes $i$ and $j$ are connected to form an edge $(i,j)$ whenever the node $i$ is the parent node for node $j$. For a node $i$ in the system, the associated three-phase  is denoted by $\rho_{i}, \phi_{i} \in \lbrace a,b,c \rbrace $. At node $i \in \mathcal{N}$ for a phase $\rho$, the complex voltage and apparent power of the load are denoted by $V_i^{\rho}$ and $S_{Li}^{\rho}=P_{Li}^{\rho}+j Q_{Li}^{\rho}$, respectively. For an edge $(i,j) \in \mathcal{E}$ and for phase $\rho$, the apparent power flow and complex current flow are shown by $S_{ij}^{\rho}$ and $I_{ij}^{\rho}$, respectively. The linear three-phase distribution power flow model is presented in equations (11)-(13). The active and reactive power balance equations are presented in equations (11) and (12), where $k: j \rightarrow k$ denotes a child node $k$ for a parent node $j$, and (13) represents linearized three-phase voltage drop equation. The variables here are active power flow, $P_{ij}^{\rho\rho}$, reactive power flow, $Q_{ij}^{\rho\rho}$, and the square of voltage magnitude, $V_{i}^{\rho}$. Although this model approximates losses, the work in \cite{vvarjha} validates that the linearized model accurately estimates nodal voltages and the power flow solutions are close to those obtained from the actual non-linear three-phase power flow model. Note that all the variables in this formulation (power flows, voltages, and current) are time-dependent; the time index is omitted for simplicity.  

\vspace{-0.4cm}
\begin{small}
\begin{eqnarray}
    P_{ij}^{\rho \rho} - p_{L,j}^{\rho} &=& \sum_{k; j \rightarrow k} P_{jk}^{\rho \rho} \hspace{0.5 cm}  \rho \in \lbrace a,b,c \rbrace\\    
    Q_{ij}^{\rho \rho} - q_{L,j}^{\rho} &=& \sum_{k; j \rightarrow k} Q_{jk}^{\rho \rho}  \hspace{0.5 cm}  \rho \in  \lbrace a,b,c \rbrace \\
    v_{i}^{\rho} - v_{j}^{\rho} &=& \sum_{q \in \phi_{j}} 2\mathbb{R}[S_{ij}^{pq} (z_{ij}^{pq})^*]   \hspace{0.5 cm}  p,q \in  \lbrace a,b,c \rbrace
\end{eqnarray}
\end{small}
\vspace{-0.3cm}

The operating constraints for bus voltages are specified in (14), where, the allowable limits are as per the ANSI Std. \cite{ANSI}. 
\begin{equation}
\small
     (V_{min})^2 \leq v_{i}^{\rho} \leq (v_{max})^2
\end{equation}

\subsection{Battery Dispatch Problem as a Markov Decision Process}
In this section, we formulate the DNO's problem as a Markov Decision Process (MDP). MDP is defined by a tuple of state space $S$, an action space $A$, a reward function $R$, a transition probability function $P_r$, and a discount factor $\gamma \in (0, 1)$. Here, the DNO is denoted as an agent for which we want to learn the optimal control algorithm for battery dispatch. The agent and distribution grid interact at each of a sequence of discrete time steps $t = \{1,2,3,..,T\}$.  At each time step $t$, the agent observes the state of the environment $s_t \in S$ and selects a control action $a_t \in A$. One time step later, the agent receives a numerical reward $R(s_t, a_t, s_{t+1}) \in R$ and a numerical cost $R^c(s_t, a_t, s_{t+1}) \in R$ and finds itself in a new state $s_{t+1}$. The states, actions, and rewards for the constrained MDP model used in this work are defined as below.

\subsubsection{State Space}
Let $e_{ESS}^t$ denote the SoC of the BESS at time step $t$, and $E_{ESS}$ represent the maximum capacity of the BESS. The energy capacity constraint of the BESS, with the single-trip charge/discharge efficiency, denoted by $\eta$, is considered
as follows:
\begin{equation}
\small
\eta E_{ESS} \leq e_{ESS}^t \leq (1-\eta)E_{ESS}
\end{equation}
The net power demand ($P_{net}$) is the input provided by the TSO and is distributed to the individual BESS as per (16)
\begin{equation}
\small
P_{net} = \sum_{i=1}^{m} p_{i}
\end{equation}

\noindent where $p_i$ is the power level charge/discharge of the BESS; $m$ is the number of available BESS at that discrete interval; and $P, Q, t$ are the nodal real, reactive power injections, and the time-interval, respectively. Taking all the above in consideration, the state, $s_t$, of the DNO agent (centralized controller) at time step $t$ is given by $s_t = [e_{ESS}^t, P_{net}^t, P^t, Q^t, V^t] \in S $.

\subsubsection {Action Space}
To satisfy $P_{net}^t$ at each time step $t$, the aggregator should choose one action from the action space $A$ for each BESS. Let $a_t \in A_{s_t}$ be the action taken at time step $t$, where $A_{s_t}$ denotes the possible action set in the action space $A$ under state $s_t$. At each time step $t$, $A_{s_t}$ is constrained by the BESS capacity, meaning that at time step $t$, only actions that satisfy the SoC condition of the BESS in the next time step $t + \tau$ given by (11) can be included in the action set, $A_{s_t}$. Here, the policy is parameterized along with the value function. The action set consists of real numbers, with actions chosen from a normal (Gaussian) distribution. Specifically, the sampled actions are re-parameterized with $\hat{a}_ \theta = \mu_{\theta} + v_{\theta} \mathscr{N} (0, 1)$, where $\mu_{\theta}$ and $v_{\theta}$ are the outputs of mean values and variances from the Gaussian policy network. $\mathscr{N} (0, 1)$ is the standard normal distribution. Therefore, $\hat{a}_ \theta$ is differentiable with respect to $\theta$.



\subsubsection{Reward Function}
The reward function $r(s_t, a_t)$ in this case returns a cost value of operating the BESS by the DNO. The DNO's operating profit is calculated as the market revenue minus the cost of aging for the BESS detailed in (17). 
\vspace{-0.1cm}
\begin{equation}
\small
max \sum_{i=1}^m  \mathscr{P}_{i} (Cr+b) \delta C - A(b) 
\end{equation}
\vspace{-0.3cm}

\noindent where, $\mathscr{P}(.)$ is the performance index calculation function (see (10)), $C$ is the regulation capacity, $r$ is the set of regulation instructions, $b$ is the set of battery power dispatch, $\delta$ is the market clearing price, and $A(.)$ is the cycle aging cost function.

To account for the impact of the current action on future rewards, the total discounted reward at time step $t$ under a given policy $\pi$, denoted by $R_t$, is defined as the sum of the instant reward at time step $t$ and discounted rewards from the next time step, $t + \tau$, given by (18).
\begin{equation}
\small
R_t =  r(s_t, a_t) + \sum_{i=1}^h \gamma^i r(s_{t+i\tau}, a_{t+i\tau})
\end{equation}
\noindent where, $\gamma $ denotes the discount factor that determines the importance of future rewards from the next time step, $t + \tau$, to end of simulation, $h$. For example, $\gamma= 0$ implies that the aggregator will consider only the current reward, while $\gamma = 1$ implies that the system weighs both current reward and future long-term rewards equally. To maintain nodal voltage profiles within a desirable range, an additional cost, $R_t^c$, is imposed for the number of voltage constraint violations observed across all nodes (19). 
\begin{equation}
\small
R_t^c = \sum_{i=1}^N \left\{(|V_{i}^{t+1}| > \bar{V}) + (|V_{i}^{t+1}| > \underline{V}) \right\}
\end{equation}
\noindent where, $V_{i}^{t+1}$ is the voltage at node $i$ at time $t+1$; $\bar{V}$ and $\underline{V}$ are upper and lower limits for voltage magnitudes; and $N$ is the total number of nodes. By evaluating the feedback in the form of rewards and costs defined above via past and/or future interactions with the physical environment, the DNO agent tries to learn a control policy that minimizes the total operational cost while satisfying the voltage constraints.

\section{Methodologies}
\begin{figure}[t]
    \centering
    \includegraphics[width=0.5\textwidth]{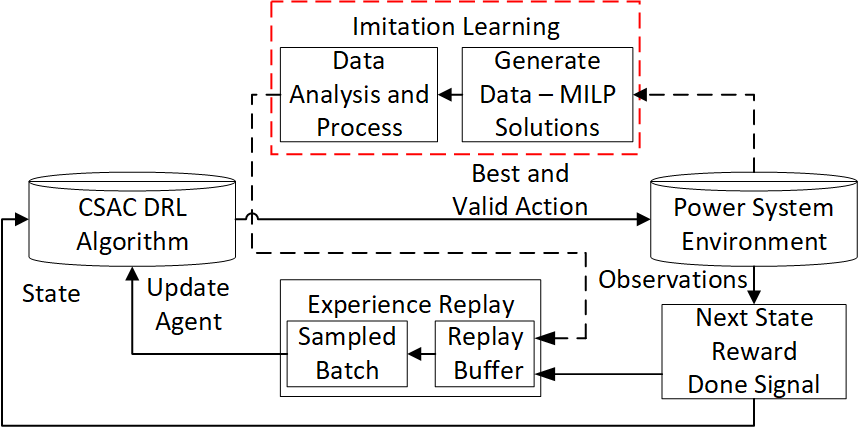}
    \caption{Overview of the DRL-IL Framework }
    \label{framework}
    \vspace{-0.5cm}
\end{figure}

In this section, we detail a learning-based controller for the DNO agent using DRL with IL improvements. The architecture of training DRL agents for the DNO's MDP problem  formulated in Section II.B is presented in Fig \ref{framework}. First, we detail the constrained soft actor critic algorithm (CSAC) to train a DRL agent for the DNO. An experience replay model is used to increase the mini-batch update efficiency. Next, we introduce soft-Q imitation learning (SQIL) algorithm to replace the memory buffer with sampled demonstrations i.e., the data generated from solving the power grid environment using OPF models. The resulting algorithm is termed as CSAC-IL for simplicity. 

CSAC is an off-policy RL agent trained in an iterative manner. In each iteration, a stochastic gradient descent is performed to update the parameters of the two sets of neural networks that approximate the state-value and action-value functions.  For the constrained optimization problem for DNO agent, we limit the number of total voltage constraints violations at each time step $R_{t}^c \leq \bar{R^c}$. The corresponding state-value function limit, $\bar{V^c},$ with the operating constraints is denoted by $V^{c, \pi} (s) \leq \bar{V^c} = (1-\gamma^T)/(1-\gamma)\bar{R^c}$, where, T is the episode length. The solution of the CSAC method adapted from \cite{RL6} is presented in Algorithm \ref{csacalgo}. The goal of the agent is to find a control policy $\pi$ that maximizes the expected discounted return while also maximizing the operating profit of the DNO subject to the distribution network voltage and participating BESS SoC constraints on the expected discounted return of the cost function.

Now, we introduce the application of IL in training the DRL algorithm for the DNO agent via using the model-based information. Specifically, we employ SQIL to improve the training of DRL agent. SQIL performs soft Q-learning (off-policy learning) with three small, but important, modifications: (1) it initially fills the agent’s experience replay buffer with demonstrations, where the rewards are set to a constant $r = +1$; (2) as the agent interacts with the world and accumulates new experiences, it adds them to the replay buffer, and sets the rewards for these new experiences to a constant $r = 0$; and (3) it balances the number of demonstration experiences and new experiences (50\% each) in each sample from the replay buffer \cite{sqil}. The SQIL approach encourages long-horizon imitation by providing the agent with (1) an incentive to imitate the demonstrated actions in demonstrated states, and (2) an incentive to take actions that lead it back to demonstrated states when it encounters new, out of distribution states  without adversarial training, by using constant rewards instead of learned rewards. The key idea is that, instead of using a learned reward function to provide a reward signal to the agent, we can simply give the agent a constant reward of $r = +1$ for matching the demonstrated action in a demonstrated state, and a constant reward of $r = 0$ for all other behavior. The SQIL framework is simple to implement with the soft actor-critic RL algorithm used in this work.  

\begin{algorithm}[t]
 \small
     \mbox{\emph{Algorithm 1: Constrained Soft Actor Critic}}\\
     \noindent\rule{8.4cm}{1pt}\\
         \textbf  {Input}: {Input Variables: time-intervals for analysis, $t$}\\
     {Initialize network parameters and lagrange multiplier $\lambda$  }\\
     {Repeat}\\
     \For {each sample step}  {
      {$a_t ~ \pi (.|s_t)$} \\
      {$D \leftarrow D \cup (s_t, a_t, s_{t+1}, R_t, R_t^c$)}
      }
      \For {each gradient step with sampled batch }  {
         {Update action value networks $Q_{\phi}$} \\
         {Update state value networks $V_{\Phi}$}\\
        {Update policy network $\pi_{\theta}$}\\
  {Update $\lambda$ }\\
     
 } 
 {Until Converge}
     \noindent\rule{8.4cm}{1pt}\\
     \label{csacalgo}
     \vspace{-10pt}
 \end{algorithm}


\section{Case Studies and Discussions}

\subsection {Simulation Setup}
We use a power grid simulator, OpenDSSCmd, to represent the environment for power distribution system, which is built upon the Pypower open-source tool for power grid simulations \cite{pypower}. The simulator is able to emulate a large-scale power distribution system with various operating conditions. The framework is developed in Python using OpenAI Gym, with an interface designed and provided for Reinforcement Learning. The RL agents are trained and tuned using python scripts allowing interactions with the OpenDSS environment. The OpenDSSCmd provides a channel for the users to visualize the system operating statuses and evaluate control actions in real time. Several power system models are available in this environment with data sets representing realistic time-series operating conditions. All market parameters and associated data sets can be directly downloaded from PJM 2016 frequency regulation market available online \cite{pjmmanual}. The DRL agents are trained using Python 3.7 scripts on a Linux server with 24 CPU cores and 64 GB of memory.

The RL agent interacts with the distribution grid every 10 ms, and each episode contains 450 time steps, for a total of $\approx$ 4 seconds per episode.  The parameter settings for the reinforcement learning algorithm is provided in Table \ref{Parameters}.  For the ACE regulation signal, we use one month of real regulation market data from PJM in March 2016, which we use to generate random samples for our simulation at the start of every episode. The data set contains 10 scenarios with regulation data for 28 continuous days. Each scenario has 21,600 time steps, each representing a 4-sec interval. Numerical simulations are conducted on IEEE 34-bus and 123-bus distribution system models available in OpenDSS which is used to solve the unbalanced power flow \cite{OpenDSS}. For the 34-bus and 123-bus distribution feeder model, 5 and 10 BESS are distributed to provide ACE regulation respectively each having a rated power capacity of $\pm 10 kW$ with 4.21 kWh energy rating. The BESS units can thus provide 50 kW and 100 kW of generations for the frequency regulation services for the two distribution feeder models respectively. The active power output and the SoC of the BESS are monitored through the OpenDSS platform. 

Based on numerical results, we found that the solution framework with the added SQIL component is more sample-efficient and trains with fewer fluctuations and variance in episodic rewards over the learning process. On the other hand, the framework with only the RL algorithm takes a bit less computation time per iteration, yet takes more iterations to converge. We also validate the constrained formulation by comparing it with the solutions obtained upon solving the standard MILP problem.

\begin{table}[t]
    \centering
    \caption{Summary of DRL training Parameters}
    \vspace{-0.3cm}
    \label{Parameters}
    \begin{tabular}{c|c|c}    
        \hline
        Hyper  & \multicolumn{2}{c}{Values}   \\
        \cline{2-3}
       Parameters & 34-bus & 123-bus \\
        \cline{2-3}
        \hline
        Hidden Layers & [64,32] & [64,32] \\
        \hline
        Activation Funtion & relu & relu \\
        \hline
        Batch Size  & 256 &    256  \\
        \hline
       Discount Factor  & 0.99 &    0.97   \\
        \hline
        Parameter Noise & 0.05 &    0.02  \\
        \hline
        Learning rate &    0.001 & 0.01   \\
        \hline
        Env Coeff & 0.1 &   0.1  \\
        \hline
    \end{tabular}
    
   \vspace{-0.5cm}
\end{table}


\subsection {Performance Evaluation of the DRL-IL Framework}

\subsubsection {Training Parameters and Results}
\begin{figure}[t]
    \centering
    \includegraphics[width=0.48\textwidth]{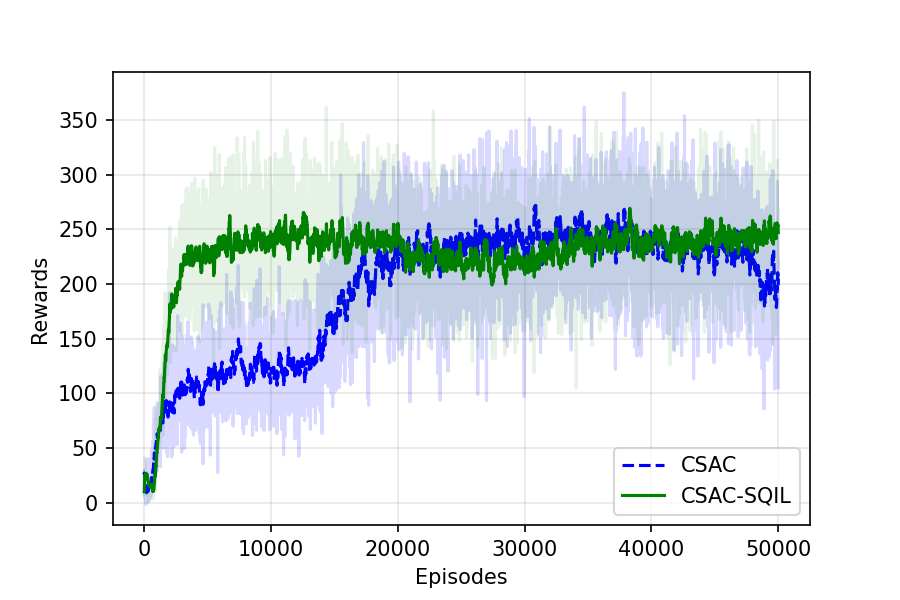}
    \vspace{-0.4cm}
    \caption{IEEE 123 Bus Training Results: Reward Functions}
    \label{train}
    \vspace{-10 pt}
\end{figure}

In this section, we evaluate the training parameters and convergence with the proposed learning methods. IEEE 34-bus and 123-bus distribution feeder models with dispersed BESS are used here for the simulations. The test system model is first trained only using the CSAC RL algorithm for 50,000 episodes, with 450 time steps for each episode, starting with a random scenario and a random sample drawn from the training data set. The training parameters are listed in Table \ref{Parameters}. For this simulation using IEEE 123-bus system, the training results are presented in Figure \ref{train}. As can be seen, the training improves over the increasing number of episodes and stabilizes after approx. 20,000 episodes. Conventionally during training, the experiences are collected and stored in the experience/replay buffer. However, here we use a sampling technique which uses a memory buffer of sampled values instead of all experiences as can be seen in Figure \ref{framework}. For the next simulation, we add an imitation learning component presented in Figure \ref{framework}. Here, the agent's experience replay buffer is updated based on the SQIL method described in Section III.C. The demonstrations are obtained from solving the same problem using MILP formulation as presented in Section II.A. The training results for this simulation is shown in Figure \ref{train} and compared with the CSAC trained agent. As it can be seen from this figure, just by penalizing the agent to only follow the demonstrations, the training is stable after about 2000 episodes and the agent reaches the maximum stable reward within 4500 episodes.




We also compare the performance of both the 34-bus and 123-bus distribution feeder model training using the proposed learning solution framework with the standard MILP solutions. The average profits, voltage violations and the time taken for the 10 epochs from March 2016 regulation data is presented in Table \ref{comp}. As observed from the table, the learning model combined with the imitation learning component yields the results closest to the mathematical solution and is solved much faster than the standard MILP model. This is because the embedded physical model information in the replay buffer from the sampled solution batch guides the training of the learning platform closer to the optimal solution.

\begin{table}[t]
    \centering
    \caption{Comparison of RL training results with MILP}
    \vspace{-0.3cm}
    \label{comp}
    \begin{tabular}{c|c|c|c|c|c}    
        \hline
        Algorithm   & \multicolumn{2}{c|}{Average}  & \multicolumn{2}{c|}{Average}  & Average   \\
        & \multicolumn{2}{c|}{Profit(k\$)} & \multicolumn{2}{c|}{Voltage Violation} & Computation \\
        \cline{2-5}
        & 34-bus & 123-bus & 34-bus & 123-bus & Time (s) \\
        \hline
        CSAC & 82.21 & 190.77 &  0.26 &  0.34 & 0.155 \\
        \hline
        CSAC-SQIL & 98.44 & 197.65 & 0.08 & 0.19 & 0.179 \\
        \hline
       MILP  & 102.19 & 207.6 0  & 0 & 0 & ~12 \\
        \hline
    \end{tabular}
   \vspace{-15pt}
\end{table}

\subsubsection {Testing Parameters and Results}
\begin{figure}[t]
    \centering
    \includegraphics[width=0.5\textwidth]{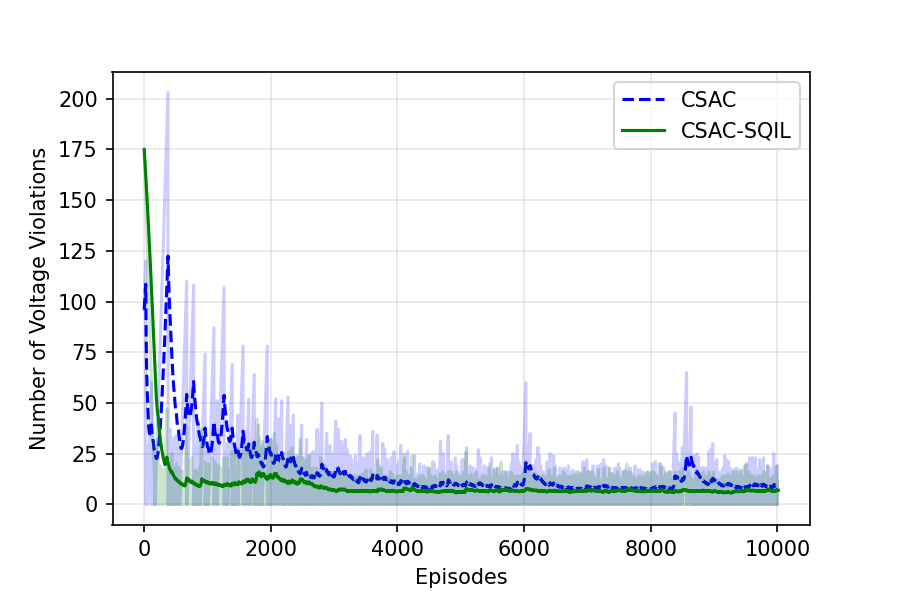}
    \vspace{-0.8cm}
    \caption{IEEE 123 Bus Testing Results: Number of Voltage Violations Over Every Testing Episode}
    \label{Vgecomp}
    \vspace{-20 pt}
\end{figure}

In this section, we evaluate the trained agent by testing through 10,000 episodes of the unseen data set by the training agent. The overall testing results are presented in Table \ref{comp2}. We also evaluate the physical model constraints of the developed learning methods.The critical constraint for this study is the voltage violation limits. We are dispatching the BESS to follow the regulation instruction while maximizing profits for the aggregator and simultaneously maintaining the voltage limit constraints of the distribution feeder model. For this study, we use the IEEE 123-node distribution feeder model with specified lower and higher voltage limits.  The part of the reward function corresponding to the voltage violation costs are observed during the testing process with the CSAC trained agent and CSAL-SQIL trained agent and the results are presented in Figure \ref{Vgecomp}. As can be seen from the figure, the number of voltage violations are significantly reduced and maintained by the CSAC-SQIL trained agent within the first 400 episodes of the testing scenario. 

\begin{table}[t]
    \centering
    \caption{Testing Results of the Trained Agent}
    \vspace{-0.3cm}
    \label{comp2}
    \begin{tabular}{c|c|c|c|c}    
        \hline
        Algorithm   & \multicolumn{2}{c|}{Average}  & \multicolumn{2}{c}{Average}    \\
        & \multicolumn{2}{c|}{Profit(k\$)} & \multicolumn{2}{c}{Voltage Violation}  \\
        \cline{2-5}
        & 34-bus & 123-bus & 34-bus & 123-bus \\
        \hline
        CSAC & 62.54 & 171.17 &  0.14 &  0.47  \\
        \hline
        CSAC-SQIL & 69.99 & 175.64 & 0.07 & 0.15 \\
        \hline
       MILP  & 77.28 & 183.56  & 0 & 0  \\
        \hline
    \end{tabular}
   \vspace{-20pt}
\end{table}

\section{Conclusion}
We presented a novel learning-based controller for optimal battery energy storage dispatch by a DNO to meet pre-specified power demand target obtained from TSO while maintaining the distribution-level operating constraints. Specifically, we proposed a constrained soft actor-critic algorithm (CSAC) for the DRL model, and a variant of behavioral cloning method called soft-Q imitation learning, CSAC-SQIL, to improve the sample efficiency and performance of the DRL agent. The proposed IL algorithm incorporates the physics-based optimal solutions as example demonstration trajectories to train the DRL agent. It is shown that with the embedded physical information using the proposed imitation leaning improvements to DRL algorithms, we are able to match the results provided by baseline mathematical optimization methods. The learning methods are compared against a MILP based optimization formulation and several system performance metrics have been evaluated. It is shown that the learning-based methods can provide faster solutions (almost 100x times) compared to the baseline mathematical optimizers while appropriately maintaining the critical network-level constraints for the power distribution grid. We observe that the performance of these DRL algorithms combined with imitation learning model is significantly improved because the neural networks are enriched with the power system model information available from the baseline optimization methods.

\bibliographystyle{ieeetr}
\vspace{-10pt}
\bibliography{references}

\end{document}